\documentclass{article}
\usepackage[preprint]{corl_2026} 

\usepackage[utf8]{inputenc}
\usepackage[T1]{fontenc}
\usepackage{amsmath,amssymb}
\usepackage{graphicx}
\usepackage{booktabs}
\usepackage{multirow}
\usepackage{enumitem}
\usepackage{siunitx}
\usepackage{xspace}
\usepackage{microtype}
\usepackage[font=small,labelfont=bf]{caption}

\newcommand{\sbf}[1]{\textbf{#1}\xspace}

\title{Neural Fidelity Calibration:\\Turning Field Anomalies into Simulation Fine-Tuning}

\author{Youwei Yu \quad Lantao Liu\\
Indiana University, Bloomington, IN, USA \quad \texttt{\{youwyu, lantao\}@iu.edu}}

\begin{document}
\maketitle

\begin{abstract}
A policy that transfers from simulation to a lab robot is not yet a policy that survives the field. Calibrating a simulator from real data usually targets a few hand-picked parameters, and even then simulators miss real effects while environments reconstructed from onboard perception inherit sensing errors. We propose Neural Fidelity Calibration (NFC), which uses conditional flow matching to infer online, from a robot's own execution, both simulator physical parameters and \emph{residual fidelity}: the dynamics discrepancy that remains after calibration and the gap between perceived and true environment. A simulation-trained anomaly detector decides from onboard signals alone when fine-tuning is needed, and a hallucination policy modulates posterior uncertainty under optimism in the face of uncertainty. NFC gives the most accurate posteriors among established simulation-based inference methods on five robots with up to \num{39} parameters and a $128\times128$ residual terrain, and on a wheeled robot with a broken wheel axle in snow, mud, and wet grass it raises rough-terrain success from \SI{26}{\percent} to \SI{70}{\percent} and reduces orientation jerk by \SI{88}{\percent}.
\end{abstract}

\keywords{Sim-to-real adaptation, simulation-based inference, flow matching, field robots}

\section{Introduction}
Zero-shot sim-to-real RL transfers agile policies to soft~\citep{tiboni23rfdroid}, wheeled~\citep{yu2024adaptive}, aerial~\citep{Geles-RSS-24}, and legged~\citep{Joonho2024wheelleg} robots, with domain randomization (DR)~\citep{tobin2017domain} as the standard tool for robustness. In the field, however, DR ranges set by experts~\citep{ma2024dreureka} and hand-reconstructed environments~\citep{escontrela2024learning} do not anticipate the operating distribution, and rare events such as hardware damage decide success. The sim-to-real-to-sim paradigm~\citep{stachowicz2024lifelong} improves the policy with real data while keeping risky exploration in simulation; within it, simulation-based inference (SBI)~\citep{papamak16efree} narrows DR by inferring simulator parameters from real trajectories. Two obstacles remain: SBI assumes a well-calibrated simulator, yet simulators omit real effects and environments built from onboard sensors inherit perception errors; and kernel- and optimization-based estimators~\citep{ramos2019bayessim, sobanbabu2025sampling} target a handful of parameters and do not scale to the dimensionality met in deployment.

This paper addresses two of the workshop's questions. For \emph{lab metric to field guarantee}, we detect transfer failure from onboard signals alone with a detector trained purely in simulation; for \emph{adaptation in the field}, we fine-tune the deployed policy online against a simulator calibrated to what the robot just experienced. Our contributions are:
\begin{enumerate}[leftmargin=*, label=(\arabic*), itemsep=0pt, topsep=2pt]
\item \textbf{Simulator-to-reality shift.} We separate \emph{simulator calibration} (parameters aligning simulated with real dynamics) from \emph{residual fidelity}, what remains after correct calibration plus the error of environments reconstructed from uncertain perception.
\item \textbf{Neural Fidelity Calibration (NFC).} A conditional flow matching estimator infers both jointly in a latent space, scaling to high-dimensional parameters and rich conditioning (state--action histories plus elevation or depth maps).
\item \textbf{Informative adaptation.} The policy is fine-tuned only when an anomaly signals imminent failure, a hallucination policy modulates NFC uncertainty following optimism in the face of uncertainty~\citep{JMLR:v11:jaksch10a, curi2020efficient}, and all real data is reused offline.
\end{enumerate}
Prior adaptive-DR methods~\citep{ramos2019bayessim, pmlr-v164-muratore22a, memmel2024asid, Heiden21neuralsim} assume a simulator that can, in principle, mirror reality; we model the residual explicitly instead. Safe-RL guarantees~\citep{hsuzen2022sim2lab2real, Shi-RSS-24} hold relative to an assumed model and disturbance set, which an unmodeled fault can violate, motivating continual recalibration in the field.

\begin{figure}[t]
\centering
\includegraphics[width=0.74\textwidth]{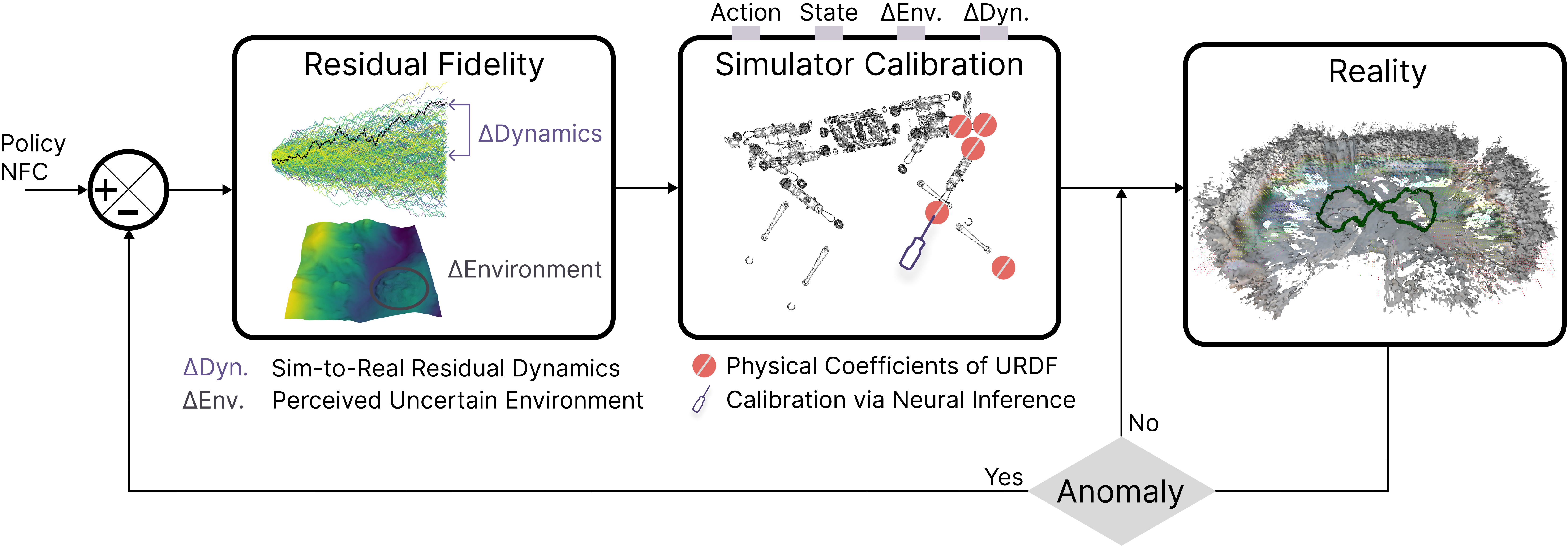}
\vspace*{-0.05in}
\caption{NFC-driven adaptation. (a)~\textbf{Residual fidelity}: colored trajectories are calibrated-simulator rollouts, black is the real execution, their difference the residual dynamics; the circled terrain region marks uncertain perception, where NFC samples residual environments to reconstruct terrain variants. (b)~\textbf{Simulator calibration}: flow matching infers physical parameters matching the real trajectory. (c)~\textbf{Informative adaptation}: policy and NFC are pretrained in simulation and fine-tuned only when an anomaly is detected.}
\label{fig:system}
\end{figure}

\section{Neural Fidelity Calibration}
We model a contextual MDP~\citep{hallak2015cmdp} whose context $c$ comprises simulator parameters $\boldsymbol{\psi}$ and environment variables $\boldsymbol{e}$ (e.g., terrain geometry), and maximize return averaged over the posterior of contexts consistent with the real execution, $\max_{\boldsymbol{\pi}} \mathbb{E}_{c\sim p(c\vert\boldsymbol{\tau}^{\mathrm{real}})}\mathbb{E}_{\boldsymbol{\tau}\sim P^{\boldsymbol{\pi}}_{c}}[R(\boldsymbol{\tau})]$. Calibrating to a single best-fit $\boldsymbol{\psi}^{*}$ instead overfits, since no single setting explains the data when the simulator is imperfect and real data is scarce. Fig.~\ref{fig:system} summarizes the framework.

\paragraph{Calibration and fidelity shifts.}
BayesSim-style inference~\citep{ramos2019bayessim} explains the real trajectory only through the simulator's own parameterization, i.e., it assumes $p(\boldsymbol{\psi}\vert\boldsymbol{\tau}^{\mathrm{real}})\approx p(\boldsymbol{\psi}\vert\boldsymbol{\tau}^{\mathrm{sim}})$ for some $\boldsymbol{\psi}$. This fails when the simulator omits real effects and when it is calibrated against a terrain perceived with error~\cite{yu2025neural}. We therefore decompose real-world transitions as
\begin{equation}\label{eqn:sys-dynamics}
s_{t+1} = \underbrace{f_{\boldsymbol{\psi}}(s_t,a_t\vert\boldsymbol{e})}_{\mathrm{Calibration}} + \underbrace{g_{\boldsymbol{\phi}}(s_t,a_t\vert\boldsymbol{e})}_{\mathrm{Fidelity}} + \epsilon(s_t,a_t\vert\boldsymbol{e}),
\end{equation}
where $f_{\boldsymbol{\psi}}$ is the nominal simulator (mass, friction, \dots), $g_{\boldsymbol{\phi}}$ models residual effects, and $\epsilon$ is noise. The residual fidelity $\boldsymbol{\phi}=\{\Delta s_{t+1},\Delta\boldsymbol{e}\}$ comprises the \emph{residual dynamics} $\Delta s_{t+1}$ (sim-vs-real state difference under identical conditions) and the \emph{residual environment} $\Delta\boldsymbol{e}$ (perceived vs.\ true environment). For a wheeled robot, $\boldsymbol{\psi}\in\mathbb{R}^{13}$ holds body and wheel masses, wheel damping, and friction, and $\boldsymbol{\phi}$ holds residual dynamics over pose, velocity, and previous action ($\mathbb{R}^{15}$) and a residual elevation map $\Delta\boldsymbol{e}\in\mathbb{R}^{128\times128}$. We assume compact supports with bounded densities (e.g., at most two broken wheels on a four-wheeled robot); NFC needs only these bounds, not real-world anomaly labels or a fault taxonomy.

We learn the joint posterior $p(\boldsymbol{\psi},\boldsymbol{\phi}\vert\boldsymbol{\tau})$ directly with conditional flow matching~\citep{lipman2022flow, wildberger2023flow}: a velocity field $\boldsymbol{v}_{\boldsymbol{\theta}}([\boldsymbol{\psi}(k),\boldsymbol{\phi}(k)],\boldsymbol{\tau},k)$ is regressed on simulator-generated parameter--trajectory pairs, and sampling integrates it over $k\in[0,1]$ from a Gaussian with a few Euler steps, which makes online use practical. We write $\boldsymbol{q}_{\boldsymbol{\theta}}$ for the induced conditional density; $N$ posterior samples reconstruct $N$ environments $\boldsymbol{e}=\hat{\boldsymbol{e}}+\Delta\boldsymbol{e}$ that reset the $N$ parallel simulators used for fine-tuning.

\paragraph{Sequential and latent NFC.}
At deployment the simulation-pretrained estimator serves as proposal prior, $\tilde{p} = \boldsymbol{q}^{\mathrm{sim}}(\cdot\vert\boldsymbol{\tau}^{\mathrm{real}})$, concentrating online simulation on plausible parameters; an estimator trained on proposal-generated data converges to the true posterior reweighted by $\tilde{p}/p$~\citep{papamak16efree}, which we use directly. To make the $128\times128$ residual environment tractable, an encoder $\mathcal{H}$ (a causal TCN~\citep{BaiTCN2018} plus an MLP for the perception map) compresses the execution data. In simulation both the noisy trajectory $\hat{\boldsymbol{\tau}}$ the robot would observe and the clean, privileged trajectory $\boldsymbol{\tau}$ are available, and their embedding difference defines the \emph{latent residual fidelity} $\boldsymbol{\phi}_z = \mathcal{H}(\boldsymbol{\tau}) - \mathcal{H}(\hat{\boldsymbol{\tau}})$. The estimator conditions on $\mathcal{H}(\hat{\boldsymbol{\tau}})$ and infers $\boldsymbol{q}(\boldsymbol{\psi},\boldsymbol{\phi}_z\vert\cdot)$; a decoder $\mathcal{Z}$ maps $\boldsymbol{\phi}_z$ back to $\boldsymbol{\phi}$. All three are trained jointly on the negative log-density plus a reconstruction loss $\|\mathcal{Z}(\boldsymbol{\phi}_z)-\boldsymbol{\phi}\|_2^2$, akin to privileged-information distillation~\citep{he2024bridging}.

\paragraph{Hallucinated randomness.}
When the posterior is highly uncertain, sampling it directly can stall fine-tuning: very low friction makes the robot slide, very high friction makes it stick, and reward is scarce in a short online window. A hallucination policy $\boldsymbol{\pi}_{\mathfrak{h}}$ that never acts on the robot controls how far each fine-tuning domain may deviate from the posterior center. With $\boldsymbol{x}=[\boldsymbol{\psi},\boldsymbol{\phi}_z]\in\mathbb{R}^{D}$ a joint latent sample, $\boldsymbol{\mu}$ the mean of the $N$ samples in the current round, and a gate $\boldsymbol{g}\sim\boldsymbol{\pi}_{\mathfrak{h}}(\cdot\mid\mathcal{H}(\hat{\boldsymbol{\tau}}))$, $\boldsymbol{g}\in(0,1]^{D}$, per parallel environment, each simulated domain is
\begin{equation}\label{eqn:hallucinate-domain}
\tilde{\boldsymbol{x}} = \boldsymbol{\mu} + \boldsymbol{g}\odot(\boldsymbol{x}-\boldsymbol{\mu}), \qquad \boldsymbol{x}\sim\boldsymbol{q}(\cdot\mid\hat{\boldsymbol{\tau}}),
\end{equation}
so $\boldsymbol{g}=\boldsymbol{1}$ recovers the posterior and $\boldsymbol{g}\to\boldsymbol{0}$ collapses to its mean, per coordinate. Control and hallucination policies are optimized jointly with PPO~\citep{schulman2017proximal} to maximize $\mathbb{E}_{\boldsymbol{x}\sim\boldsymbol{q}}\mathbb{E}_{\boldsymbol{g}\sim\boldsymbol{\pi}_{\mathfrak{h}}}\mathbb{E}_{\boldsymbol{\tau}\sim P^{\boldsymbol{\pi}}_{\tilde{\boldsymbol{x}}}}[R(\boldsymbol{\tau})]$.

\paragraph{Anomaly-triggered deployment.}
A nominal reference window $\boldsymbol{\tau}^{\mathfrak{c}}$ is compared with the current window $\boldsymbol{\tau}^{\mathfrak{t}}$~\citep{darban2025carla} by a classifier two-sample test~\citep{lopez2016revisiting, koch2015siamese} on the NFC TCN: the logit $\ell_{\mathrm{A}} = \boldsymbol{w}^{\top}\vert\mathcal{H}(\boldsymbol{\tau}^{\mathfrak{c}})-\mathcal{H}(\boldsymbol{\tau}^{\mathfrak{t}})\vert + b$ is trained with binary cross-entropy on simulated pairs only, nominal when physical and residual parameters are retained and anomalous when the target's parameters are redrawn from the prior. A window is flagged when $\ell_{\mathrm{A}}$ exceeds the \num{99}th percentile on held-out nominal pairs. Online, each flagged window triggers one adaptation round: NFC infers the posterior from the latest $H$-step window, the hallucination policy gates its spread, gated samples reset $N$ parallel simulators, the policy is fine-tuned, and sequential NFC is updated with the gated samples as proposal. After the mission, the accumulated data refine NFC, detector, and policy.

\section{Experiments}
\paragraph{Sim-to-sim.}
We evaluate \texttt{Ant}, \texttt{Quadruped} (Go1), \texttt{Humanoid} (H1), \texttt{Quadcopter} (Ingenuity), and \texttt{Jackal} with $17$, $25$, $39$, $9$, and $13$ physical parameters, respectively, plus residual fidelity over proprioceptive state and a $128\times128$ residual elevation map (a residual depth map for \texttt{Quadcopter}). Policies and NFC are pretrained in Genesis~\citep{Genesis} and deployed in MuJoCo~\citep{todorov2012mujoco}, whose different contact and integration models plus sensing and actuation noise provide the gap, in \emph{Flat} (obstacle-free) and \emph{Rough} (randomized elevation maps~\citep{yu2024adaptive} or obstacle fields~\citep{kulkarni2023aerialgymisaac}) environments. Each online fine-tuning round takes \SI{0.2}{\second}; online NFC training (\SI{1.75}{\second}) runs in parallel with the \SI{2}{\second} execution window. We compare our flow matching (FM) head with TCN features against FM with \sbf{MLP}, \sbf{QMC-RFF}~\citep{ramos2019bayessim}, or \sbf{QMC-FKS}~\citep{johannes2024slice} features, mixture density networks \sbf{MDN-RFF} (BayesSim) and \sbf{MDN-TCN}, and \sbf{CMA-ES}~\citep{hansen2016cma, sobanbabu2025sampling}.

Measured by the average log-density of the ground-truth parameters under the predicted posterior~\citep{ramos2019bayessim}, NFC is best in all ten robot--environment settings: on \emph{Flat}/\emph{Rough} it reaches $0.57$/$0.46$ (\texttt{Ant}), $-0.42$/$-0.46$ (\texttt{Quadruped}), $-1.03$/$-2.64$ (\texttt{Humanoid}), $-0.90$/$-1.50$ (\texttt{Quadcopter}), and $1.21$/$0.69$ (\texttt{Jackal}), against $-1.00$/$-1.09$, $-1.14$/$-1.34$, $-1.63$/$-2.93$, $-1.20$/$-1.86$, and $0.26$/$-1.10$ for the strongest baseline, FM-MLP; Accuracy drops from \emph{Flat} to \emph{Rough}, where perception enters the conditioning, and as $\vert\boldsymbol{\psi}\vert$ grows; fixed random bases (RFF, FKS) cannot adapt to high-dimensional conditioning, and CMA-ES needs \num{100} rather than \num{10} simulation iterations, too slow for online use. The detector, trained only on Genesis pairs and tested on \num{20000} MuJoCo windows per setting at a threshold calibrated to \SI{99}{\percent} specificity, reaches a true-positive rate of \num{0.97} (\emph{Flat}) and \num{0.80} (\emph{Rough}) versus \num{0.85}/\num{0.65} for MLP and at most \num{0.70}/\num{0.51} for kernel features; in the field it flagged all \num{2160} anomalous trajectories. In normalized return, PPO w/ NFC beats PPO w/o $\Delta$Fidelity (physical calibration only), PPO w/o NFC (fine-tuned directly on execution data), PPO w/ MDN-RFF, PPO w/ CMA-ES, and the model-based TD-MPC2~\citep{hansen2024tdmpc} in all ten settings (e.g., \texttt{Jackal} \emph{Rough}: \num{0.82} vs.\ \num{0.69}, \num{0.20}, \num{0.52}, \num{0.50}, \num{0.41}); removing residual fidelity lowers return everywhere, by \num{0.03} (\texttt{Humanoid}) to \num{0.34} (\texttt{Quadruped}).

\begin{figure}[t]
\centering
\begin{minipage}[c]{0.36\textwidth}
\centering
\includegraphics[width=\textwidth]{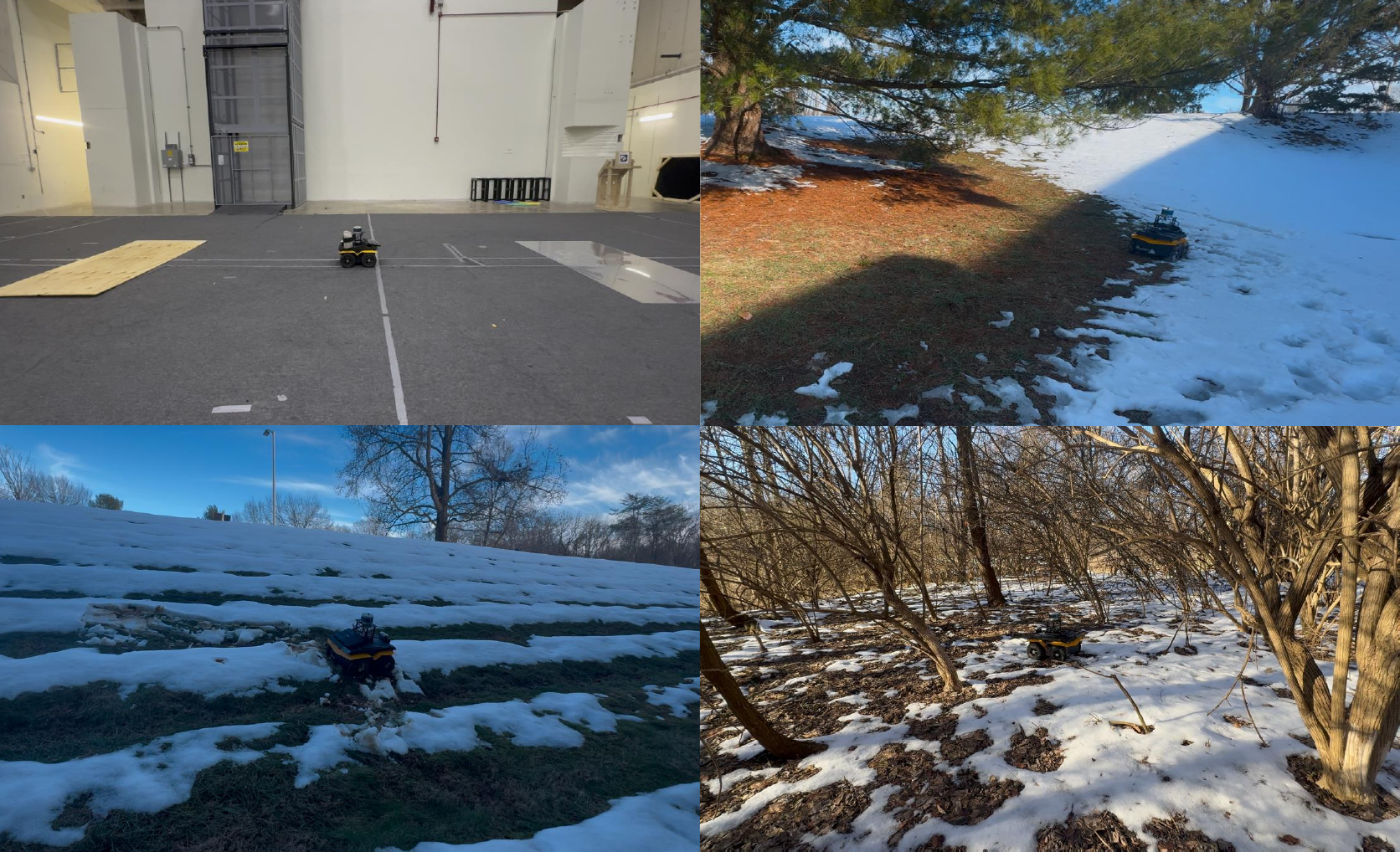}
\end{minipage}\hfill
\begin{minipage}[c]{0.60\textwidth}
\centering
\includegraphics[width=\textwidth]{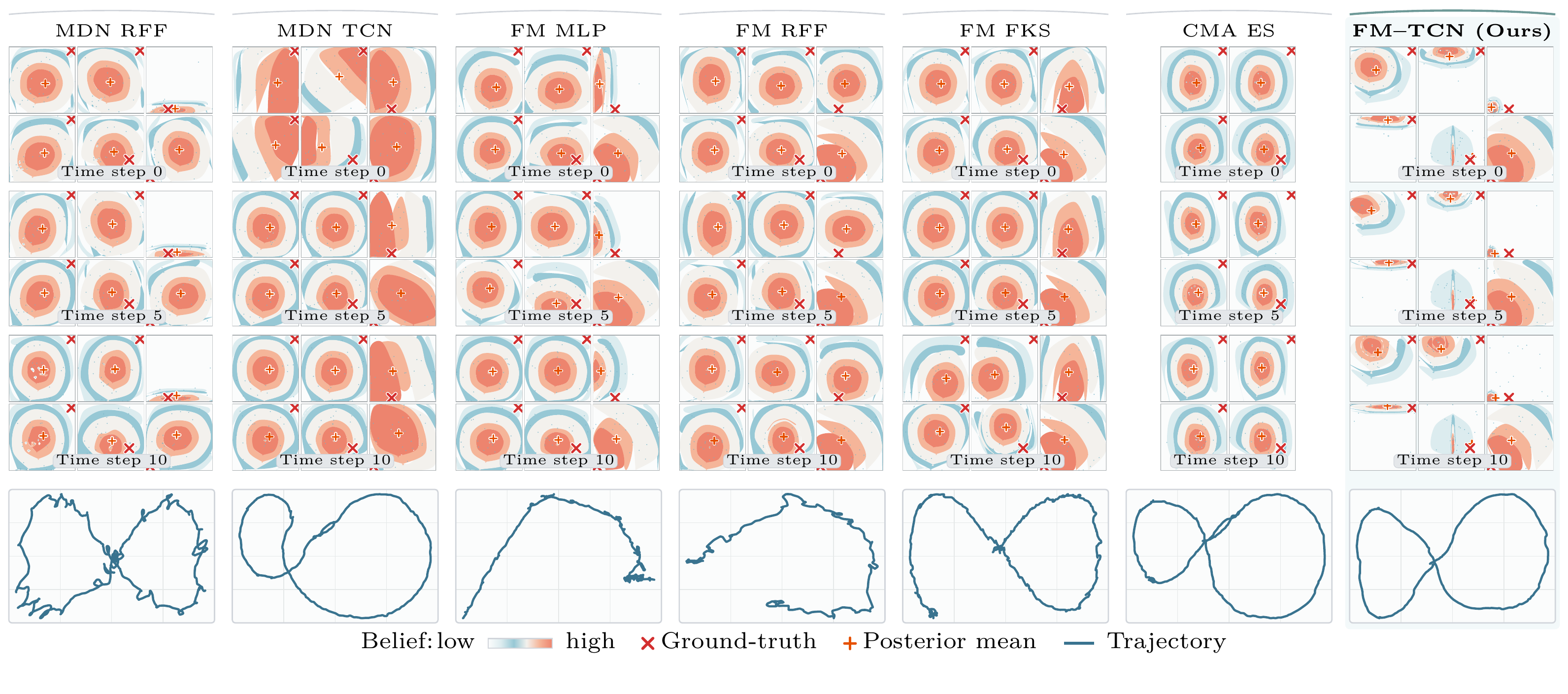}
\end{minipage}
\vspace*{-0.05in}
\caption{\textbf{Left:} field scenarios with a broken front-left wheel axle (lab surfaces with and without a loose rock; snow, wet grass, mud, and elevation change outdoors). \textbf{Right:} NFC on real snow. Per history window, each method block shows joint posteriors over wheel damping ($x$) and friction ($y$) for the four wheels (top: rear-right, rear-left; bottom: front-right, front-left) and the residual dynamics ($\Delta x,\Delta y$) and terrain height ($\Delta z$). Only NFC singles out the front-left wheel with near-zero damping.}
\label{fig:real}
\end{figure}

\paragraph{Field deployment.}
The Jackal builds elevation maps~\citep{RAL18-elemap} onboard from depth and LiDAR-inertial odometry~\cite{fasterlio,yu2023mimu}; policy optimization runs on a laptop GPU. All scenarios use a \emph{broken front-left wheel axle} (Fig.~\ref{fig:real}, left): four laboratory \emph{Flat} challenges (standard or mixed-material floor, each with and without a loosely attached rock on the robot) and three wild \emph{Rough} challenges (wet grass and snow, elevated wet grass and snow, mud and snow in a forest). The task is a \SI{3}{\meter}-diameter figure-eight of waypoints; a lap succeeds when every waypoint is reached within \SI{0.25}{\meter} without rollover or intervention (\num{9} runs per controller and scenario). Beyond the baselines above we add \sbf{Falco}~\citep{Zhang2020Falco}, a \SI{200}{\hertz} motion-primitive planner with elevation maps~\citep{cao2023granularity}, and report success rate, trajectory ratio (traveled over target path length; \num{1.0} is ideal), and orientation and position jerk~\citep{yu2024adaptive}.

\begin{table}[t]
\scriptsize
\centering
\setlength{\tabcolsep}{2pt}
\caption{Field results with a broken front-left wheel axle over four \emph{Flat} and three \emph{Rough} scenarios, \num{9} runs each.}
\label{tab:real}
\begin{tabular*}{\textwidth}{@{\extracolsep{\fill}} ll ccccccc @{}}
\toprule
& & \begin{tabular}[c]{@{}c@{}}PPO\\ w/ NFC\end{tabular} & \begin{tabular}[c]{@{}c@{}}PPO\\ w/o $\Delta$Fidelity\end{tabular} & \begin{tabular}[c]{@{}c@{}}PPO\\ w/o NFC\end{tabular} & \begin{tabular}[c]{@{}c@{}}PPO\\ w/ MDN-RFF\end{tabular} & \begin{tabular}[c]{@{}c@{}}PPO\\ w/ CMA-ES\end{tabular} & TD-MPC2 & Falco \\
\midrule
\multirow{2}{*}{Success Rate (\%) $\uparrow$}
  & \emph{Flat}  & $\mathbf{100}$ & $\mathbf{100}$ & $\mathbf{100}$ & $\mathbf{100}$ & $\mathbf{100}$ & $\mathbf{100}$ & $\mathbf{100}$ \\
  & \emph{Rough} & $\mathbf{70}$  & $52$  & $26$  & $48$  & $44$  & $41$  & $30$ \\
\multirow{2}{*}{Traj.\ Ratio $\downarrow$}
  & \emph{Flat}  & $1.80$ & $1.80$ & $1.90$ & $1.53$ & $1.99$ & $\mathbf{1.50}$ & $1.60$ \\
  & \emph{Rough} & $\mathbf{2.10}$ & $3.20$ & $3.70$ & $2.40$ & $2.35$ & $2.70$ & $3.60$ \\
\multirow{2}{*}{Orien.\ Jerk (\si{\radian\per\second\cubed}) $\downarrow$}
  & \emph{Flat}  & $\mathbf{2.98}$ & $3.50$  & $14.18$ & $6.14$ & $7.96$  & $6.95$  & $41.90$ \\
  & \emph{Rough} & $\mathbf{3.19}$ & $8.49$  & $26.99$ & $9.81$ & $10.38$ & $8.98$  & $41.93$ \\
\multirow{2}{*}{Pos.\ Jerk (\si{\meter\per\second\cubed}) $\downarrow$}
  & \emph{Flat}  & $\mathbf{4.36}$ & $9.32$  & $8.41$  & $6.92$ & $7.79$  & $5.27$  & $6.95$ \\
  & \emph{Rough} & $\mathbf{6.26}$ & $12.46$ & $12.42$ & $9.39$ & $11.44$ & $18.46$ & $8.87$ \\
\bottomrule
\end{tabular*}
\end{table}

Fig.~\ref{fig:real} (right) shows calibration on snow: the inferred damping of the front-left wheel averages \num{0}, consistent with a free-spinning wheel, and friction on all four wheels averages \num{0.4}, reflecting the snow, whereas competing methods trained on the same data produce diffuse posteriors. Ground truth for residual fidelity is unavailable in the field, so its effect is measured through policy improvement (Table~\ref{tab:real}). PPO fine-tuned only on real data (PPO w/o NFC) succeeds on \emph{Flat} but with high jerk, and on \emph{Rough} falls into repetitive back-and-forth motion. PPO w/o $\Delta$Fidelity, which injects the perceived map directly into simulation, improves over it (\SI{52}{\percent} vs.\ \SI{26}{\percent} \emph{Rough} success) but stays well below the full method; only $\Delta$Fidelity models the remaining residual dynamics and perception uncertainty. Falco's primitives ignore the asymmetric traction of the broken axle and yield the highest orientation jerk. Overall, NFC raises \emph{Rough} success from \SI{26}{\percent} to \SI{70}{\percent} while reducing orientation jerk by \SI{88}{\percent}.

\section{Limitations and Conclusion}
Detection contrasts two windows, so slowly developing changes produce little signal and are absorbed only offline. Parameters and residuals need not be uniquely identifiable from a short trajectory, so inferred values should be read with the downstream policy result. Fine-tuning updates persist across rounds, so returning to an earlier regime does not by itself restore earlier behavior. Our field evidence is one platform across seven scenarios; distributional transfer estimates across robots and sites remain open. In summary, NFC turns field anomalies into targeted simulation fine-tuning, lifting rough-terrain success with a broken axle from \SI{52}{\percent} to \SI{70}{\percent}.

\bibliography{references}

\end{document}